\documentclass[11pt]{article}
\usepackage[preprint]{acl}

\usepackage{times}
\usepackage{latexsym}
\usepackage[T1]{fontenc}
\usepackage[utf8]{inputenc}
\usepackage{microtype}
\IfFileExists{inconsolata.sty}{\usepackage{inconsolata}}{}
\usepackage{graphicx}

\usepackage{amsmath,amssymb,amsfonts,bm}
\usepackage{booktabs}
\usepackage{multirow}
\usepackage{xcolor}
\usepackage{tikz}
\usetikzlibrary{arrows.meta,positioning,fit,calc,decorations.pathreplacing}
\usepackage{pgfplots}
\pgfplotsset{compat=1.16}
\usepackage{algorithm}
\usepackage{algpseudocode}
\usepackage{float}
\usepackage{enumitem}
\setlist{nosep,leftmargin=*}

\definecolor{fixedc}{RGB}{86,132,192}
\definecolor{slidec}{RGB}{232,146,58}
\definecolor{inactc}{RGB}{222,222,222}
\definecolor{gA}{RGB}{38,140,120}
\definecolor{gB}{RGB}{150,85,175}
\definecolor{mhcc}{HTML}{8A8A8A}
\definecolor{tphcc}{HTML}{1F6FB2}
\definecolor{xhcc}{HTML}{B5562E}
\definecolor{llamac}{HTML}{2B2B2B}

\newcommand{\method}{TPHC}
\newcommand{\Hpre}{\mathcal{H}^{\mathrm{pre}}}
\newcommand{\Hpost}{\mathcal{H}^{\mathrm{post}}}
\newcommand{\Hres}{\mathcal{H}^{\mathrm{res}}}

\newcommand{\Sset}{\mathcal{I}}
\newcommand{\R}{\mathbb{R}}
\newcommand{\SK}{\mathrm{SK}}
\newcommand{\vecop}{\mathrm{vec}}
\newcommand{\mat}{\mathrm{mat}}

\title{TPHC: Improving Hyper-Connections with Tensor-Parallel\\ Grouped Updates}

\author{
Xinying Hu\thanks{Equal contribution.}\textsuperscript{1,2},
You Wu\footnotemark[1]\textsuperscript{1,2,3},
Haoyi Wu\footnotemark[1]\textsuperscript{1,2},
Jianchao Tan\textsuperscript{3}, 
Kefeng Zhang\textsuperscript{3},
Yuchen Xie\textsuperscript{3},
Kewei Tu\thanks{Corresponding author.}\textsuperscript{1,2} \\
\textsuperscript{1}School of Information Science and Technology, ShanghaiTech University \\
\textsuperscript{2}Shanghai Engineering Research Center of Intelligent Vision and Imaging \\
\textsuperscript{3}Meituan \\
\texttt{\{huxy2023, wuyou2024, wuhy1, tukw\}@shanghaitech.edu.cn}
}

\begin{document}
\maketitle

\begin{abstract}
Hyper-Connections (HC) expand the residual stream of a Transformer into $N$ parallel streams, and manifold-constrained HC (mHC) stabilizes them at scale.
However, mHC benefits little from more than four streams, partly because each sublayer writes a single output to all streams, limiting how differently they can be updated.
In this work, we propose \method{}, which integrates tensor-parallel (TP) shards with HC streams to promote greater independence across streams. Specifically, we route a separate mixture of the residual streams to each TP shard and keep the individual shard outputs separate during write-back, so each stream can receive a different mixture of those outputs.
This design makes the residual-stream updates more expressive
without adding parameters or FLOPs to the attention or FFN sublayers.
With native support for stream sharding under TP, \method{} requires only 9\% more training time than mHC on an 8B model when both use four streams.
Increasing the number of streams from four to sixteen raises its training time by only 19\%.
On a 1.1B model trained on 100B tokens, \method{} reduces validation loss by 0.7--1.0\% and raises average zero-shot accuracy across six commonsense reasoning benchmarks by 0.3--0.7 percentage points relative to mHC at matched stream counts $N\in\{4,8,16\}$.
Our code is available at \url{https://github.com/wuyou2002/TPHC}.
\end{abstract}

\section{Introduction}
\label{sec:intro}

Hyper-Connections \citep[HC;][]{zhu2025hyper} have recently attracted growing interest as an effective approach to improving the performance of large language models (LLMs).
HC expands standard residual connections \citep{he2016deep,he2016identity} from a single residual path to $N$ parallel residual streams, introducing learnable control over inter-layer information flow.
Specifically, each sublayer reads a learned mixture of the streams and writes its output back to them, while a learned $N{\times}N$ matrix mixes the residual streams.
Building on HC, Manifold-Constrained Hyper-Connections \citep[mHC;][]{xie2025mhc}
constrain this residual mixing matrix to be doubly stochastic to
stabilize signal propagation across layers.
The DeepSeek-V4 series adopts mHC \citep{deepseek2026v4}, and
DeepSeek-V4.1-Flash uses a Single-Pass variant designed to enable
more efficient kernel fusion \citep{deepseek2026v41flash}.

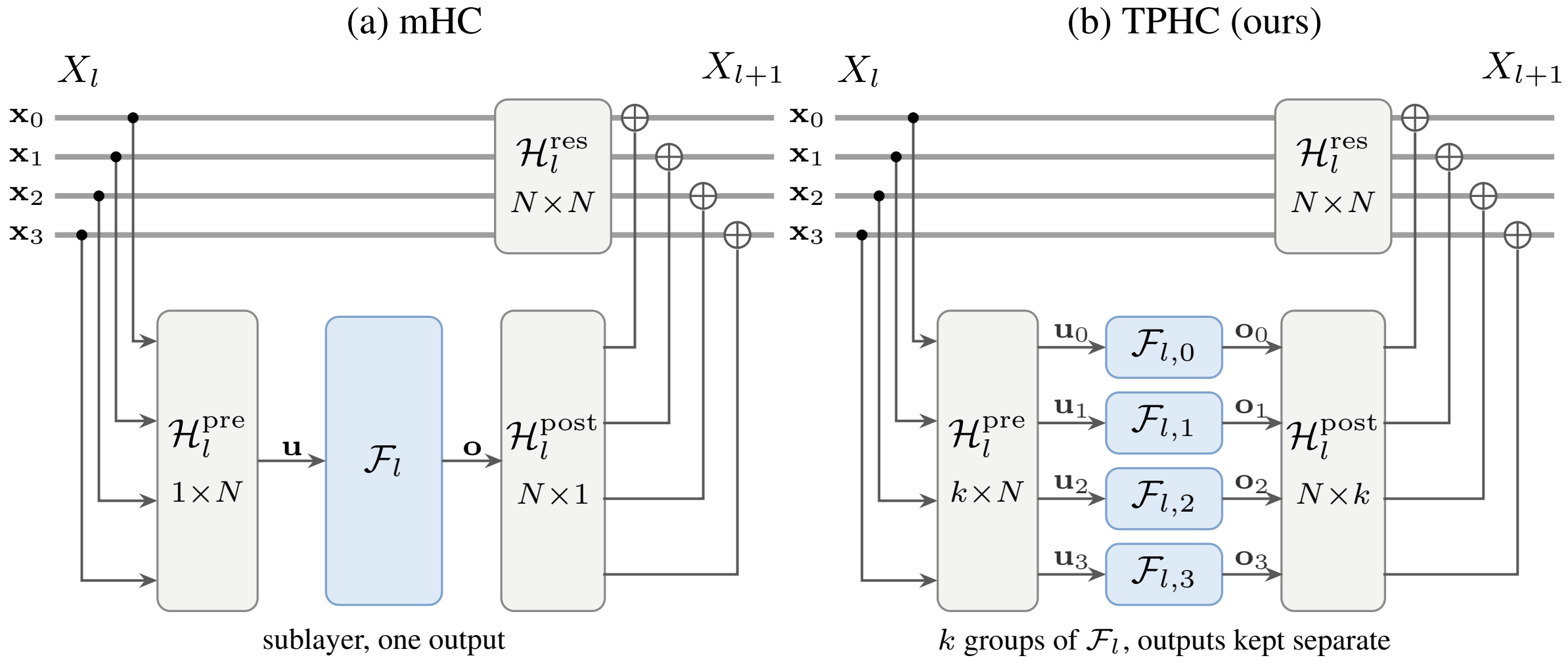
\begin{figure*}[t]
  \centering
  \resizebox{\textwidth}{!}{
\definecolor{gpuAl}{HTML}{595959}\colorlet{gpuAf}{white}\definecolor{gpuBl}{HTML}{595959}\colorlet{gpuBf}{white}\definecolor{gpuCl}{HTML}{595959}\colorlet{gpuCf}{white}\definecolor{gpuDl}{HTML}{595959}\colorlet{gpuDf}{white}\definecolor{softblue}{HTML}{DFEAF7}\definecolor{softbluebar}{HTML}{C9DCF1}\definecolor{blueedge}{HTML}{7FA2CC}\definecolor{mapfill}{HTML}{F3F3F2}\definecolor{edge}{HTML}{8A8A8A}\definecolor{commfill}{HTML}{DCEFDF}\definecolor{commedge}{HTML}{7DB38A}%
\begin{tikzpicture}[>={Stealth[length=1.5mm, width=1.2mm]},
  lab/.style={font=\scriptsize, inner sep=1pt},
  lanemhc/.style={line width=1.3pt, draw=black!38},
  lanefix/.style={line width=1.3pt, draw=black!38},
  flow/.style={->, line width=0.6pt, draw=black!65},
  flow-/.style={line width=0.6pt, draw=black!65},
  wline/.style={line width=0.6pt, draw=black!65},
  mapbox/.style={draw=edge, line width=0.5pt, fill=mapfill, rounded corners=3pt, inner sep=1pt, font=\footnotesize},
  resbox/.style={draw=edge, line width=0.5pt, fill=mapfill, rounded corners=3pt, inner sep=1pt, align=center, font=\footnotesize},
  shard/.style={draw=edge, line width=0.5pt, fill=white, rounded corners=3pt, minimum height=0.5cm, inner sep=1pt, font=\footnotesize},
  fbox/.style={draw=blueedge, line width=0.5pt, fill=softblue, rounded corners=3pt, minimum height=0.5cm, inner sep=1pt, font=\footnotesize},
  fgray/.style={draw=edge, line width=0.5pt, fill=black!10, rounded corners=3pt, minimum height=0.5cm, inner sep=1pt, font=\footnotesize},
  commbox/.style={draw=commedge, line width=0.5pt, fill=commfill, rounded corners=3pt, inner sep=1pt, font=\footnotesize},
  coef/.style={->, densely dashed, line width=0.5pt, draw=black!50},
  arbox/.style={draw=commedge, line width=0.5pt, fill=commfill, rounded corners=3pt, inner sep=0pt, font=\scriptsize},
  addnode/.style={circle, draw=black!65, fill=white, inner sep=0pt, minimum size=0.21cm, line width=0.55pt,
                  path picture={\draw[line width=0.5pt, black!65] (path picture bounding box.north) -- (path picture bounding box.south)
                  (path picture bounding box.west) -- (path picture bounding box.east);}},
]
\node[font=\small, anchor=south] at (2.945,4.240) {(a) mHC};
\draw[lanemhc] (0.000,3.740) -- (5.890,3.740);
\node[lab, anchor=east] at (-0.030,3.740) {$\mathbf{x}_{0}$};
\draw[lanemhc] (0.000,3.420) -- (5.890,3.420);
\node[lab, anchor=east] at (-0.030,3.420) {$\mathbf{x}_{1}$};
\draw[lanemhc] (0.000,3.100) -- (5.890,3.100);
\node[lab, anchor=east] at (-0.030,3.100) {$\mathbf{x}_{2}$};
\draw[lanemhc] (0.000,2.780) -- (5.890,2.780);
\node[lab, anchor=east] at (-0.030,2.780) {$\mathbf{x}_{3}$};
\node[font=\footnotesize, anchor=south] at (0.200,3.860) {$X_l$};
\node[font=\footnotesize, anchor=south] at (5.640,3.860) {$X_{l+1}$};
\draw[flow] (0.640,3.740) -- (0.640,1.910) -- (0.840,1.910);
\fill (0.640,3.740) circle (0.045);
\draw[flow] (0.500,3.420) -- (0.500,1.257) -- (0.840,1.257);
\fill (0.500,3.420) circle (0.045);
\draw[flow] (0.360,3.100) -- (0.360,0.603) -- (0.840,0.603);
\fill (0.360,3.100) circle (0.045);
\draw[flow] (0.220,2.780) -- (0.220,-0.050) -- (0.840,-0.050);
\fill (0.220,2.780) circle (0.045);
\node[mapbox, align=center, minimum width=0.82cm, minimum height=2.46cm] at (1.250,0.930) {$\mathcal{H}^{\mathrm{pre}}_l$\\[3pt]{\scriptsize$1{\times}N$}};
\node[fbox, minimum width=0.95cm, minimum height=2.36cm] at (2.695,0.930) {$\mathcal{F}_{l}$};
\node[lab, anchor=north] at (2.695,-0.400) {sublayer, one output};
\node[mapbox, align=center, minimum width=0.82cm, minimum height=2.46cm] at (4.080,0.930) {$\mathcal{H}^{\mathrm{post}}_l$\\[3pt]{\scriptsize$N{\times}1$}};
\draw[flow] (1.660,0.930) -- node[lab, above] {$\mathbf{u}$} (2.220,0.930);
\draw[flow] (3.170,0.930) -- node[lab, above] {$\mathbf{o}$} (3.670,0.930);
\node[resbox, minimum width=0.95cm, minimum height=1.26cm] at (4.080,3.260) {$\mathcal{H}^{\mathrm{res}}_l$\\[1pt]{\scriptsize$N{\times}N$}};
\draw[wline] (4.490,1.860) -- (4.750,1.860) -- (4.750,3.620);
\node[addnode] at (4.750,3.740) {};
\draw[wline] (4.490,1.240) -- (5.030,1.240) -- (5.030,3.300);
\node[addnode] at (5.030,3.420) {};
\draw[wline] (4.490,0.620) -- (5.310,0.620) -- (5.310,2.980);
\node[addnode] at (5.310,3.100) {};
\draw[wline] (4.490,0.000) -- (5.590,0.000) -- (5.590,2.660);
\node[addnode] at (5.590,2.780) {};
\node[font=\small, anchor=south] at (9.335,4.240) {(b) \method{} (ours)};
\draw[lanemhc] (6.390,3.740) -- (12.280,3.740);
\node[lab, anchor=east] at (6.360,3.740) {$\mathbf{x}_{0}$};
\draw[lanemhc] (6.390,3.420) -- (12.280,3.420);
\node[lab, anchor=east] at (6.360,3.420) {$\mathbf{x}_{1}$};
\draw[lanemhc] (6.390,3.100) -- (12.280,3.100);
\node[lab, anchor=east] at (6.360,3.100) {$\mathbf{x}_{2}$};
\draw[lanemhc] (6.390,2.780) -- (12.280,2.780);
\node[lab, anchor=east] at (6.360,2.780) {$\mathbf{x}_{3}$};
\node[font=\footnotesize, anchor=south] at (6.590,3.860) {$X_l$};
\node[font=\footnotesize, anchor=south] at (12.030,3.860) {$X_{l+1}$};
\draw[flow] (7.030,3.740) -- (7.030,1.910) -- (7.230,1.910);
\fill (7.030,3.740) circle (0.045);
\draw[flow] (6.890,3.420) -- (6.890,1.257) -- (7.230,1.257);
\fill (6.890,3.420) circle (0.045);
\draw[flow] (6.750,3.100) -- (6.750,0.603) -- (7.230,0.603);
\fill (6.750,3.100) circle (0.045);
\draw[flow] (6.610,2.780) -- (6.610,-0.050) -- (7.230,-0.050);
\fill (6.610,2.780) circle (0.045);
\node[mapbox, align=center, minimum width=0.82cm, minimum height=2.46cm] at (7.640,0.930) {$\mathcal{H}^{\mathrm{pre}}_l$\\[3pt]{\scriptsize$k{\times}N$}};
\node[fbox, minimum width=0.95cm] at (9.085,1.860) {$\mathcal{F}_{l,0}$};
\node[fbox, minimum width=0.95cm] at (9.085,1.240) {$\mathcal{F}_{l,1}$};
\node[fbox, minimum width=0.95cm] at (9.085,0.620) {$\mathcal{F}_{l,2}$};
\node[fbox, minimum width=0.95cm] at (9.085,0.000) {$\mathcal{F}_{l,3}$};
\node[lab, anchor=north] at (9.085,-0.400) {$k$ groups of $\mathcal{F}_l$, outputs kept separate};
\node[mapbox, align=center, minimum width=0.82cm, minimum height=2.46cm] at (10.470,0.930) {$\mathcal{H}^{\mathrm{post}}_l$\\[3pt]{\scriptsize$N{\times}k$}};
\draw[flow, draw=gpuAl] (8.050,1.860) -- node[lab, above, text=black!80] {$\mathbf{u}_{0}$} (8.610,1.860);
\draw[flow, draw=gpuAl] (9.560,1.860) -- node[lab, above, text=black!80] {$\mathbf{o}_{0}$} (10.060,1.860);
\draw[flow, draw=gpuBl] (8.050,1.240) -- node[lab, above, text=black!80] {$\mathbf{u}_{1}$} (8.610,1.240);
\draw[flow, draw=gpuBl] (9.560,1.240) -- node[lab, above, text=black!80] {$\mathbf{o}_{1}$} (10.060,1.240);
\draw[flow, draw=gpuCl] (8.050,0.620) -- node[lab, above, text=black!80] {$\mathbf{u}_{2}$} (8.610,0.620);
\draw[flow, draw=gpuCl] (9.560,0.620) -- node[lab, above, text=black!80] {$\mathbf{o}_{2}$} (10.060,0.620);
\draw[flow, draw=gpuDl] (8.050,0.000) -- node[lab, above, text=black!80] {$\mathbf{u}_{3}$} (8.610,0.000);
\draw[flow, draw=gpuDl] (9.560,0.000) -- node[lab, above, text=black!80] {$\mathbf{o}_{3}$} (10.060,0.000);
\node[resbox, minimum width=0.95cm, minimum height=1.26cm] at (10.470,3.260) {$\mathcal{H}^{\mathrm{res}}_l$\\[1pt]{\scriptsize$N{\times}N$}};
\draw[wline] (10.880,1.860) -- (11.140,1.860) -- (11.140,3.620);
\node[addnode] at (11.140,3.740) {};
\draw[wline] (10.880,1.240) -- (11.420,1.240) -- (11.420,3.300);
\node[addnode] at (11.420,3.420) {};
\draw[wline] (10.880,0.620) -- (11.700,0.620) -- (11.700,2.980);
\node[addnode] at (11.700,3.100) {};
\draw[wline] (10.880,0.000) -- (11.980,0.000) -- (11.980,2.660);
\node[addnode] at (11.980,2.780) {};
\end{tikzpicture}}
  \caption{One sublayer (attention or FFN) of (a) mHC and (b) \method{} with $N{=}4$ streams. Horizontal lines are residual streams.
  (a) In mHC, the sublayer reads one mixture $\mathbf{u}$ of the streams, and its single output $\mathbf{o}$ is written to every stream, while $\Hres_l$ mixes all $N$ streams densely.
  (b) In \method{}, the sublayer is split into $k{=}4$ groups: group $j$ reads its own mixture $\mathbf{u}_j$ of the streams and returns its own output $\mathbf{o}_j$, and $\Hpost_l$ gives each stream its own mixture of the $k$ outputs. The illustrated case shows the full update, which writes all $N$ streams. The sparse update (\S\ref{sec:sparse}) restricts $\Hpost_l$ and $\Hres_l$ to the $k$ active streams.}
  \label{fig:arch}
\end{figure*}

Most HC-family models use $N{=}4$, since additional streams bring little gain while substantially increasing computational cost.
xHC \citep{zhang2026xhc} attributes these limited gains to the shared source of stream updates: each stream receives a scaled copy of the same sublayer output.
Therefore, it proposes \emph{temporal feature augmentation} (TFA) to address this problem, which derives additional write-back components from causal convolutions of the FFN outputs from nearby preceding tokens.

In this work, we explore the possibility of using the partial outputs of tensor-parallel shards as a complementary source of write-back diversity.
Large models commonly use tensor parallelism \citep[TP;][]{shoeybi2019megatron} for training and inference.
Consider the computation graph of mHC with TP sharding.
TP splits each attention and FFN sublayer across $k$ devices by heads or hidden units, so that each shard computes a partial output from a shared input.
An all-reduce aggregates these partial outputs into a single output, which mHC then writes to all streams.
Thus, the $k$ partial outputs already computed by TP are collapsed into the single write-back component.
Instead, we propose \textbf{\method{}} (\textbf{T}ensor-\textbf{P}arallel grouped \textbf{H}yper-\textbf{C}onnections), which treats the shards as groups and preserves their outputs as separate write-back components (Figure~\ref{fig:arch}).
In this way, each sublayer contributes $k$ separate write-back components rather than one.
In addition, each group reads its own learned mixture of the $N$ streams, allowing group-specific inputs as well as outputs.
Unlike xHC, our grouped write-back applies to both attention and FFN sublayers by adapting the existing HC read and write
mappings to the TP groups, without introducing additional
feature-augmentation modules.
\method{} increases the training time of mHC by only 7--8\% on 1.1B and 8B models when both use four streams.

This grouped design also naturally supports tensor parallelism.
Each group resides on a separate device, naturally partitioning the group inputs and outputs across devices.
The residual streams can also be sharded accordingly, with both read and write operations implemented using reduce-scatter collectives.
This reduces the memory cost of the residual streams and makes training models with larger stream counts feasible: at the 8B scale, increasing $N$ from $4$ to $16$ raises the training time of \method{} by only 19\%, compared with 59\% without sharding.


In order to further reduce the cost of the mappings, we adopt the sparse update of xHC \citep{zhang2026xhc} with a fixed schedule, so each layer writes only $k$ of the $N$ streams while every sublayer still reads all of them.
This reduces the cost of generating the mappings from $O(N^3C)$ to $O(kN^2C+k^3C)$.
In addition, we initialize each group to read from and write to its own stream.
This group-specific initialization breaks the symmetry \citep{alimaskina2026stream} within each sublayer
and we empirically find it benefical to model performance.

Our experiments demonstrate the effectiveness of \method{}.
On a 1.1B model, \method{} achieves lower training and validation loss than mHC with $N$ from $4$ to $16$.
It also improves average zero-shot accuracy on commonsense reasoning tasks by up to 0.7 percentage points over mHC.
Notably, \method{} with 8 streams outperforms mHC with 16 streams.
Our analysis also suggests that the shared write drives the mHC streams toward a common direction, whereas the \method{} streams remain less homogeneous.

\section{Related Work}
\label{sec:related}

\subsection{Hyper-Connections: HC, mHC, and xHC}
\label{sec:related-hc}

Each Transformer block consists of an attention sublayer followed by an FFN sublayer.
We index the sublayers by $l\in\{1,2,\dots,2L\}$, where $L$ is the number of Transformer blocks.
We denote the residual state after sublayer $l$ as
$X_l=(\mathbf{x}_{l,0},\dots,\mathbf{x}_{l,N-1})^{\top}\in\R^{N\times C}$,
where $N$ is the number of streams and $C$ is the hidden size.
Before the first sublayer, we form the initial state $X_0$ by replicating the token embedding across all $N$ streams.
HC \citep{zhu2025hyper} replaces the residual update
$\mathbf{x}_{l}=\mathbf{x}_{l-1}+
\mathcal{F}(\mathbf{x}_{l-1},\mathcal{W}_l)$ with
\begin{equation}
  X_l=\Hres_l X_{l-1}+
  \Hpost_l\,\mathcal{F}\big(\Hpre_l X_{l-1},\mathcal{W}_l\big),
  \label{eq:hc}
\end{equation}
where $\mathcal{F}(\cdot,\mathcal{W}_l)$ is the attention or FFN sublayer with parameters $\mathcal{W}_l$.
The pre-mapping $\Hpre_l\in\R^{1\times N}$ forms the sublayer input as a weighted sum of the streams, the post-mapping $\Hpost_l\in\R^{N\times1}$ writes the output back to the streams, and the residual mapping $\Hres_l\in\R^{N\times N}$ mixes the streams.
mHC \citep{xie2025mhc} predicts these three mappings from the input state normalized by RMSNorm \citep{zhang2019rmsnorm}:
\begin{equation}
  \mathbf{x}'_l
  = \mathrm{RMSNorm}(\vecop(X_{l-1}))
  \in \R^{1\times NC}.
  \label{eq:gen}
\end{equation}
The three mappings are then parameterized as
\begin{align}
  \Hpre_l
  &= \sigma\big(
     \alpha^{\mathrm{pre}}_l
     \mat(\mathbf{x}'_l\boldsymbol{\varphi}^{\mathrm{pre}}_l)
     + \mathbf{b}^{\mathrm{pre}}_l
     \big),
     \label{eq:mhc-pre}\\
  \Hpost_l
  &= 2\sigma\big(
     \alpha^{\mathrm{post}}_l
     \mat(\mathbf{x}'_l\boldsymbol{\varphi}^{\mathrm{post}}_l)
     + \mathbf{b}^{\mathrm{post}}_l
     \big),
     \label{eq:mhc-post}\\
  \Hres_l
  &= \SK\big(\exp(
     \alpha^{\mathrm{res}}_l
     \mat(\mathbf{x}'_l\boldsymbol{\varphi}^{\mathrm{res}}_l)
     + \mathbf{b}^{\mathrm{res}}_l
     )\big).
     \label{eq:mhc-res}
\end{align}
Here $\boldsymbol{\varphi}_l$ are learned projections,
$\alpha_l$ scalar gates, $\mathbf{b}_l$ static biases and
$\sigma$ the sigmoid, and Sinkhorn--Knopp normalization
($\SK$; \citealp{sinkhorn1967concerning}) makes $\Hres_l$
doubly stochastic, which keeps the residual path non-expansive
across depth.

xHC \citep{zhang2026xhc} shares our motivation of making better use of multiple residual streams. Its \emph{sparse residual-stream architecture} mixes and writes only $k$ of the $N$ streams, a few fixed ones and others chosen per token by a learned router, while every sublayer still reads all $N$. It also adds write-back components to each FFN through TFA. Our \method{} retains the dense read and the sparse update of xHC but selects the written streams with a fixed schedule and obtains the additional write-back components from the TP partition of each sublayer.

\citet{alimaskina2026stream} trace the under-use of HC streams to their permutation symmetry at initialization and break it with Learned Stream Scaling (LSS), learnable per-stream scales applied where the residual is expanded into streams. In \method{}, we break this symmetry within each sublayer by
initializing each group to read primarily from and write primarily to a distinct stream.

mHC-lite \citep{yang2026mhclite} builds exactly doubly stochastic matrices without Sinkhorn iterations, and Frac-Connections \citep{zhu2025frac} split the hidden state into fractions instead of copying it.
Other work improves cross-layer information flow by aggregating
earlier activations instead of expanding the residual state
\citep{huang2017densely,pagliardini2024denseformer,xiao2025muddformer,kimi2026attention}.
\subsection{Tensor Parallelism}
\label{sec:related-tp}

Megatron-style TP splits each attention sublayer by heads and each FFN by its intermediate dimension across $t$ devices \citep{shoeybi2019megatron,narayanan2021efficient}.
The first projection is \emph{column-parallel}: each device computes the queries, keys and values of its heads, or its slice of the FFN units, from the replicated input.
The output projection is \emph{row-parallel}: each device produces a partial output in $\R^{C}$, and one all-reduce per sublayer sums them.
Sequence parallelism \citep{korthikanti2023reducing} rewrites each all-reduce as a reduce-scatter and an all-gather.
These collectives are a major cost, especially in latency-bound inference \citep{pope2023efficiently}.
Ladder-Residual \citep{zhang2025ladder} reroutes residual connections so that communication overlaps with computation, and SPD \citep{kim2025spd} drops synchronization points after selected attention sublayers.
\method{} likewise modifies the architecture around the TP collectives, but instead of rescheduling the all-reduce, it removes the summation of the shard outputs.

\section{Method}
\label{sec:method}

Compared with mHC, \method{} changes how each sublayer reads from and writes to the $N$ streams.
Its central component is the grouped hyper-connection, which keeps the outputs of the TP groups separate (\S\ref{sec:tphc}).
It is complemented by the sparse update (\S\ref{sec:sparse}) and an initialization that places each group on its own stream (\S\ref{sec:init}).
We describe the TP implementation, in which both groups and streams are sharded across devices (\S\ref{sec:tp}).
Algorithm~\ref{alg:tphc} in Appendix~\ref{app:alg} gives the full procedure for one sublayer.

\subsection{Tensor-Parallel Grouped Hyper-Connections}
\label{sec:tphc}

We split each sublayer into $k$ groups along its tensor-parallel partition, which divides its parameters into $\mathcal{W}_l=(\mathcal{W}_{l,0},\dots,\mathcal{W}_{l,k-1})$, and write $\mathcal{F}_{l,j}(\cdot)=\mathcal{F}(\cdot,\mathcal{W}_{l,j})$ for group $j$.
Group $j$ contains $n_h/k$ of the $n_h$ attention heads and $n_{kv}/k$ of the $n_{kv}$ KV heads in each attention sublayer, and $D/k$ of the $D$ intermediate units in each FFN sublayer, with the matching columns of the input projections and rows of the output projection. This partitioning preserves the total parameter count and
FLOPs of the attention and FFN projections.
Given a common input $\mathbf{u}$, the sum of the group outputs equals the output of the original sublayer, $\mathcal{F}(\mathbf{u},\mathcal{W}_l)=\sum_j\mathcal{F}_{l,j}(\mathbf{u})$, and this sum is the all-reduce of standard TP.
\method{} gives each group $j$ its own input $\mathbf{u}_j$ and keeps the outputs separate:
\begin{equation}
  \mathbf{o}_j=\mathcal{F}_{l,j}(\mathbf{u}_j).
  \label{eq:group}
\end{equation}

Every group reads a learned mixture of all $N$ streams, and the $k$ group outputs $O_l=(\mathbf{o}_0,\dots,\mathbf{o}_{k-1})^{\top}$ are written back to the streams:
\begin{align}
  U_l &= \Hpre_l\,X_{l-1}, \label{eq:read}\\
  X_l &= \Hres_l\,X_{l-1}+\Hpost_l\,O_l, \label{eq:fullwrite}
\end{align}
where $U_l=(\mathbf{u}_0,\dots,\mathbf{u}_{k-1})^{\top}$, $\Hpre_l\in\R^{k\times N}$, $\Hpost_l\in\R^{N\times k}$ and $\Hres_l\in\R^{N\times N}$.
In Eq.~\eqref{eq:hc}, every stream receives a multiple of the same output, whereas $\Hpost_l O_l$ here can have rank $k$.

We follow the mHC parameterization in
Eqs.~\eqref{eq:mhc-pre}--\eqref{eq:mhc-res}, adjusting the
projection dimensions and reshapes to the matrix shapes above.
For $\Hpost_l$, we use $\sigma$ rather than $2\sigma$:
\begin{equation}
  \Hpost_l = \sigma\big(
    \alpha^{\mathrm{post}}_l \mat(\mathbf{x}'_l\boldsymbol{\varphi}^{\mathrm{post}}_l)
    + \mathbf{b}^{\mathrm{post}}_l
  \big).
  \label{eq:post}
\end{equation}
Attention and FFN sublayers have separate mappings. The token embedding is copied to all $N$ streams, and the final hidden state is $\mathrm{RMSNorm}(\sum_i\mathbf{x}_{2L,i})$.

\subsection{Sparse Update}
\label{sec:sparse}

Eq.~\eqref{eq:fullwrite} updates all $N$ streams at every sublayer, which we call a \emph{full update}. Generating the three mappings from the $NC$-dimensional input state costs $6(2kN^2{+}N^3)C$ training FLOPs per token and sublayer, compared with $6(N^3{+}2N^2)C$ in mHC.

To reduce the computational cost of the mapping generation, we follow xHC to update only $k$ streams per sublayer, one per group, which we call a \emph{sparse update}.
Let $\Sset_l=(s_0,\dots,s_{k-1})$ denote the indices of these
\emph{active streams}.
The update in Eq.~\eqref{eq:fullwrite} becomes
\begin{align}
  X_l[\Sset_l] &= \Hres_l\,X_{l-1}[\Sset_l]+\Hpost_l\,O_l,
    \label{eq:write}\\
  \mathbf{x}_{l,i} &= \mathbf{x}_{l-1,i}
    \quad\text{for } i\notin\Sset_l,
    \label{eq:identity}
\end{align}
where $X_{l-1}[\Sset_l]\in\R^{k\times C}$ stacks the active streams from the input state. The mappings $\Hpost_l,\Hres_l\in\R^{k\times k}$ are generated as in Eq.~\eqref{eq:gen} from the entries
$\mathbf{x}'_{l,\Sset}\in\R^{1\times kC}$ of $\mathbf{x}'_l$
corresponding to the active streams. Each sublayer still reads all $N$ streams as in Eq.~\eqref{eq:read}.

We choose the active streams with a heuristic schedule shared by both sublayers of each Transformer block. Let $b(l)=\lfloor(l-1)/2\rfloor$ be the zero-based block index of sublayer $l$. Following xHC, we update the first two streams, the \emph{fixed} streams, at every sublayer. The remaining $r=k-2$ slots advance by $r$ positions per block over the other $N-2$ streams, which we call the \emph{sliding} streams:
\begin{align}
  \Sset_l ={}& \{0,1\}\cup
  \big\{2+(b(l)r+j)\bmod(N-2):
  \notag\\
  &\qquad j=0,\dots,r-1\big\}.
  \label{eq:schedule}
\end{align}
When $N=k$, all streams are active in every block. For $k=4$, each sliding stream is active once every 3 blocks when $N=8$ and once every 7 blocks when $N=16$, as shown in Figure~\ref{fig:schedule} in Appendix~\ref{app:schedule}.

In contrast to the router of xHC, which selects streams per token, our schedule depends only on the layer index.
It introduces no parameters
and uses fixed gather and scatter indices.
The ablation experiments show that the performance of our sparse update matches that of the full update (\S\ref{sec:ablation}).

Under the sparse update, generating the mappings only costs $6(kN^2{+}2k^3)C$ training FLOPs per token and sublayer, which is $6912C$ when $N{=}16$ and $k{=}4$, compared with $27648C$ in mHC.
Appendix~\ref{app:cost} gives the detailed analysis for training FLOPs and parameter overhead.

\subsection{Initialization}
\label{sec:init}

At initialization, the input-dependent component of each mapping
is zero, so the mapping is determined solely by its static bias.
In mHC, the token embedding is copied to all streams, and each
sublayer writes its single output to every stream with the same
weight, so the streams remain identical and the model starts as
a standard Transformer. If \method{} used the same initialization, it would write the sum of the group outputs to
every stream, so all groups would read the same input and start
as shards of a single standard sublayer.

In \method{}, we set the bias of $\Hpre_l$ in row $j$, column $s_j$ to $+3$ and all other entries in that row to $-3$. For $\Hpost_l$ and $\Hres_l$, we set the diagonal bias entries to $+3$ and the off-diagonal entries to $-3$. After applying the sigmoid to $\Hpre_l$ and $\Hpost_l$ and Sinkhorn normalization to the exponentiated $\Hres_l$, row $j$ of $\Hpre_l$ is nearly one-hot at stream $s_j$, and $\Hpost_l$ and $\Hres_l$ are close to the identity. Thus, group $j$ reads primarily from its own active stream $s_j$, and stream $s_j$ receives mainly the output $\mathbf{o}_j$ (Figure~\ref{fig:init}d). We refer to this write-back initialization as the \emph{identity write}.

The groups receive identical inputs in the first sublayer because all streams start from the same embedding. Their separate writes allow the streams to differ after that sublayer, so the groups can read different inputs in subsequent sublayers. In our ablation, identity write lowers the final training loss relative to dense write at $N=8$ and $N=16$ (\S\ref{sec:ablation}). Appendix~\ref{app:impl} gives the implementation details.

\subsection{Tensor-Parallel Implementation}
\label{sec:tp}

\begin{figure}[t]
  \centering
  \resizebox{\columnwidth}{!}{\input{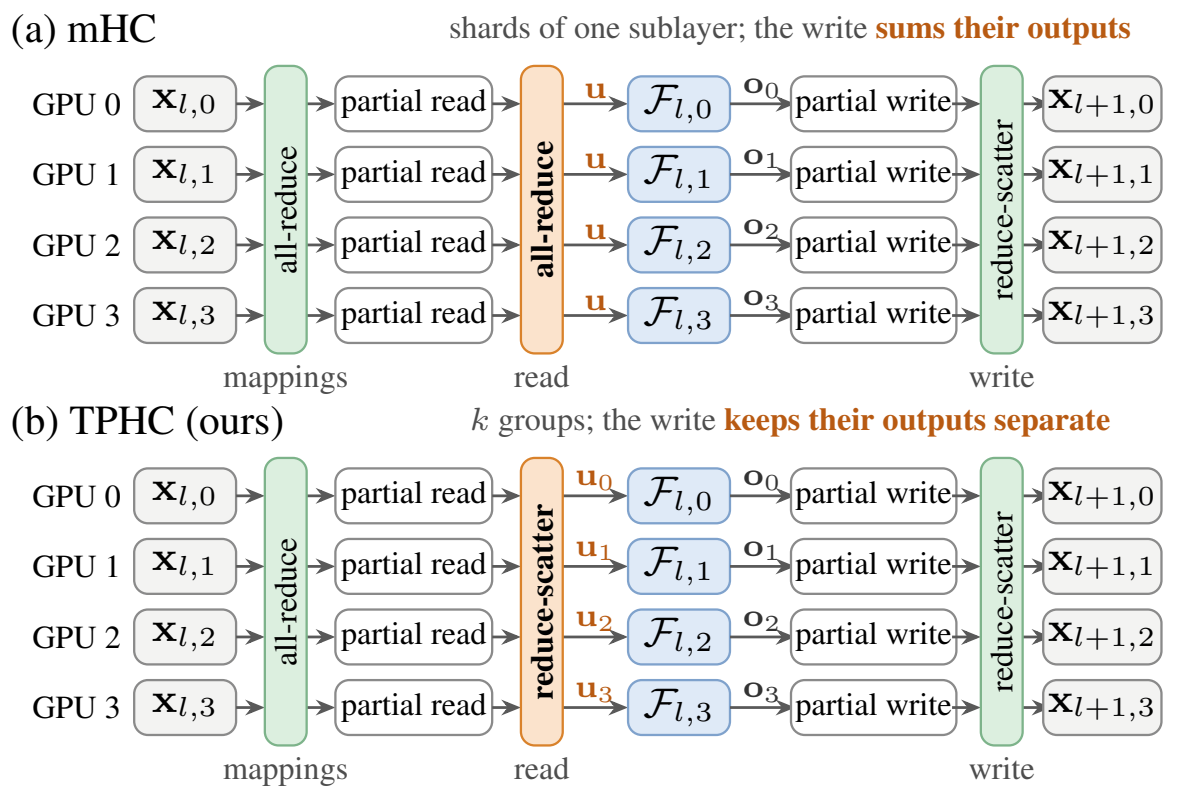}}
  \caption{One sublayer of (a) mHC and (b) \method{} under tensor parallelism of degree 4 with sharded streams. GPU $j$ holds stream $j$ and shard $j$ of the sublayer, bars are collectives, and orange marks where (a) and (b) differ. mHC all-reduces the read, since every shard needs the full input $\mathbf{u}$, and its write sums the partial outputs $\mathbf{o}_j$. \method{} reduce-scatters the read into the group inputs $\mathbf{u}_j$ and writes the $\mathbf{o}_j$ as separate components.}
  \label{fig:infra}
\end{figure}

\paragraph{Replicated streams.}

Under standard TP, mHC follows the same strategy as a standard Transformer. Every device keeps a copy of all $N$ streams and performs mapping generation, reading, and writing locally. With TP degree $t$, one all-reduce per sublayer sums the partial outputs of the shards. In this all-reduce, each device sends $2(t{-}1)C/t$ elements per token per sublayer, or $1.5C$ at $t{=}4$.

\method{} can adopt the same strategy. For $t{=}k$, device $j$ holds group $j$, which coincides with its Megatron shard. For $t{<}k$, each device holds $k/t$ groups. Each device computes its group inputs locally. Since the $k$ group outputs must remain separate, an all-gather replaces the all-reduce. In this all-gather, each device sends $(t{-}1)kC/t$ elements per token per sublayer, or $3C$ at $t{=}k{=}4$.

With replicated streams, every device stores all $N$ streams and generates all mappings, so the memory per device grows with $N$.
\paragraph{Sharded streams.}
To reduce per-device memory, we shard the residual streams so that each device holds $N/t$ of them (Figure~\ref{fig:infra}).
For mHC (Figure~\ref{fig:infra}a), each sublayer then requires three collectives.
The mappings depend on the normalized state of all streams (Eq.~\eqref{eq:gen}), so each device computes partial logits and the RMSNorm statistic from its own streams, and an all-reduce sums them.
For the read, each device applies the columns of $\Hpre_l$ that belong to its streams, and since every shard requires the full input $\mathbf{u}$, a second all-reduce sums these partial inputs.
For the write, each device combines its own streams with its partial output $\mathbf{o}_j$, and a reduce-scatter sums these contributions and returns the updated streams, so that the residual mixing, the write-back and the all-reduce of standard TP share a single collective.
At $N{=}4$, sharding raises the communication of mHC from $1.5C$ to $4.5C$ per sublayer (Table~\ref{tab:mem} in Appendix~\ref{app:cost}), and since mHC writes all $N$ streams, its write grows with $N$ ($13.5C$ in total at $N{=}16$).

\method{} shards its streams in the same way (Figure~\ref{fig:infra}b) and differs mainly in the read, where a reduce-scatter replaces the all-reduce and leaves $\mathbf{u}_j$ on device $j$.
Device $j$ then computes its group (Eq.~\eqref{eq:group}), including the output projection, without communication, and the write sums the contributions to $\Hres_l X_{l-1}[\Sset_l]+\Hpost_l O_l$ with a reduce-scatter, as in mHC.
Both reduce-scatters become all-gathers in the backward pass.
For $N{=}k$, device $j$ holds stream $j$.
For $N{>}k$, the streams are placed so that every device holds $k/t$ of the streams written in each layer, and since the fixed streams are written in every layer, the write reduce-scatter delivers each updated stream directly to
the device that will hold it next, without a separate collective.
Across the two reduce-scatters, each device sends a total of
$2(t{-}1)kC/t$ elements per token per sublayer, or $6C$ at
$t{=}k{=}4$ for any $N$. The all-reduce for the mappings operates on $kN{+}2k^2{+}1$ values per token.

Sharding thus raises the communication of \method{} from $3C$ to $6C$, and its overhead relative to mHC shrinks from $3C/1.5C{=}2$ to $6C/4.5C{\approx}1.3$.
The remaining $1.5C$ lies entirely in the read, as the cost of keeping $k$ separate outputs, and unlike that of mHC, the communication of \method{} does not grow with $N$.

\section{Experiments}
\label{sec:experiments}

We evaluate \method{} on the Llama \citep{touvron2023llama} model, comparing language modeling loss and zero-shot commonsense reasoning accuracy across different stream counts (\S\ref{sec:main}).
We test the prefill and training cost on a single GPU and under tensor parallelism (\S\ref{sec:efficiency}).
We conduct ablation studies to examine the contributions of grouped reads and writes, initialization, and the sparse updates (\S\ref{sec:ablation}). Finally, we analyze the similarity of residual streams and their write-back mixtures across depth to examine how \method{} affects stream diversity (\S\ref{sec:diversity}).

We compare models with standard residual connections (Llama), mHC \citep{xie2025mhc}, mHC-TFA, and \method{} with $k{=}4$ groups.
mHC-TFA augments mHC with the TFA module from xHC \citep{zhang2026xhc}, including Gram--Schmidt orthogonalization.
It differs from xHC only in that it writes all $N$ streams at every layer instead of using the sparse update.
We evaluate each multi-stream method with $N\in\{4,8,16\}$ streams and denote each configuration by appending the stream count to the method name (e.g.\ mHC16).
Our code builds on HuggingFace Transformers \citep{wolf2020transformers} and uses the scaled dot-product attention (SDPA) of PyTorch \citep{ansel2024pytorch} with \texttt{torch.compile} in \texttt{max-autotune} mode.
Detailed model and training configurations are provided in Appendix~\ref{app:hparams}.





\subsection{Main Results}
\label{sec:main}

\begin{table}[t]
  \centering
  \footnotesize
  \setlength{\tabcolsep}{2.6pt}
  \resizebox{\columnwidth}{!}{%
  \begin{tabular}{@{}lccccccccc@{}}
    \toprule
    & \multicolumn{7}{c}{Acc$\uparrow$} & \multicolumn{2}{c}{Loss$\downarrow$} \\
    \cmidrule(lr){2-8}\cmidrule(l){9-10}
    Model & HS & OBQA & WG & ARC-c & ARC-e & PIQA & \textbf{Avg.} & Train & Val \\
    \midrule
    Llama & 51.1 & 31.6 & 54.6 & 27.6 & 49.3 & 71.5 & 47.6 & 2.1167 & 2.1188 \\
    \midrule
    mHC4 & 52.2 & 33.0 & 56.6 & 27.6 & 50.4 & \textbf{72.1} & 48.6 & 2.0967 & 2.0985 \\
    mHC-TFA4 & 52.4 & \textbf{33.2} & \textbf{56.7} & 27.2 & 52.0 & 72.0 & \textbf{48.9} & 2.0904 & 2.0923 \\
    \method{}4 & \textbf{52.7} & 33.0 & 54.9 & \textbf{28.3} & \textbf{52.6} & 71.9 & \textbf{48.9} & \textbf{2.0819} & \textbf{2.0838} \\
    \midrule
    mHC8 & 52.4 & 34.0 & 55.2 & 27.8 & 50.4 & \textbf{72.8} & 48.8 & 2.0907 & 2.0924 \\
    mHC-TFA8 & 52.8 & \textbf{34.8} & 54.9 & \textbf{28.3} & \textbf{51.5} & 71.7 & 49.0 & 2.0841 & 2.0859 \\
    \method{}8 & \textbf{54.2} & \textbf{34.8} & \textbf{56.2} & 26.9 & 51.1 & 72.4 & \textbf{49.3} & \textbf{2.0689} & \textbf{2.0706} \\
    \midrule
    mHC16 & 52.8 & 34.6 & 55.9 & 28.8 & 50.5 & 71.4 & 49.0 & 2.0867 & 2.0885 \\
    mHC-TFA16 & 53.1 & 32.8 & \textbf{57.2} & 28.2 & \textbf{52.5} & \textbf{72.1} & 49.3 & 2.0818 & 2.0835 \\
    \method{}16 & \textbf{54.2} & \textbf{35.8} & 55.2 & \textbf{29.3} & 51.9 & 71.9 & \textbf{49.7} & \textbf{2.0671} & \textbf{2.0688} \\
    \bottomrule
  \end{tabular}}
  \caption{Zero-shot accuracy (\%) and final training and validation loss of 1.1B models trained on 100B tokens. The number after each name is $N$, and \method{} uses $k{=}4$. The best result for each $N$ is in bold. HS, OBQA, and WG denote HellaSwag, OpenBookQA, and WinoGrande, respectively.}
  \label{tab:main}
\end{table}

We follow the configuration of TinyLlama \citep{zhang2024tinyllama}, a Llama model with 1.1B parameters.
We train the models from scratch on a 100B-token subset of the SlimPajama \citep{soboleva2023slimpajama} dataset with sequence length 2048 and a global batch of 2M tokens.
The models are trained with the AdamW optimizer \citep{loshchilov2018decoupled} with $\beta_1{=}0.9$ and $\beta_2{=}0.95$.
We use a cosine learning rate schedule with a maximum learning rate of $4\times10^{-4}$ and a warmup of 200 steps.
The final learning rate is $4\times10^{-5}$.
We use a weight decay of 0.1 and gradient clipping of 1.0.
We report the final training loss, averaged over the last 50 steps, and the loss on the full SlimPajama validation set (163M tokens).
We also report zero-shot accuracy with the LM Evaluation Harness \citep{gao2023harness} on commonsense reasoning tasks, including HellaSwag \citep{zellers2019hellaswag}, OpenBookQA \citep{mihaylov2018openbookqa}, WinoGrande \citep{sakaguchi2020winogrande}, ARC-Challenge and ARC-Easy \citep{clark2018arc}, and PIQA \citep{bisk2020piqa}.

Table~\ref{tab:main} reports the main results.
All hyper-connection variants outperform Llama.
\method{} achieves the lowest training and validation loss at every $N$, 0.7--1.0\% below mHC and 0.4--1.7\% below mHC-TFA on both, and improves average zero-shot accuracy relative to mHC by 0.3--0.7 percentage points.

\begin{figure}[t]
  \centering
  \input{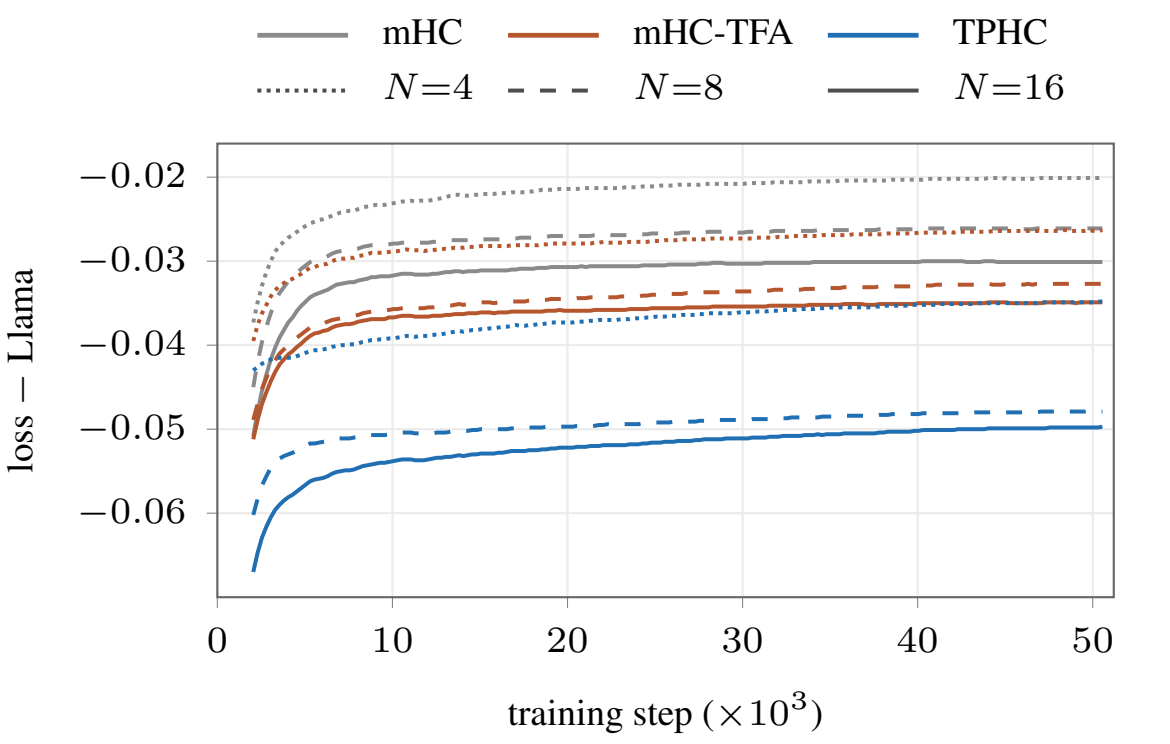}
  \caption{Training loss of the 1.1B hyper-connection models minus that of Llama, smoothed over 1k steps. Color marks the method and line style the number of streams $N$. Lower is better.}
  \label{fig:loss}
\end{figure}

\method{} also requires fewer streams to reach lower loss.
\method{}4 matches mHC-TFA16, and \method{}8 outperforms both 16-stream baselines.
As shown in Figure~\ref{fig:loss}, all methods exhibit their largest gain over Llama early in training, while the gap between \method{} and mHC emerges within the first tenth of training and remains nearly constant.

\begin{table*}[t]
  \centering
  \footnotesize
  \setlength{\tabcolsep}{8pt}
  \renewcommand{\arraystretch}{0.95}
  \begin{tabular}{@{}lrrrrrrrr@{}}
    \toprule
    & \multicolumn{4}{c}{1.1B} & \multicolumn{4}{c}{8B} \\
    \cmidrule(lr){2-5}\cmidrule(l){6-9}
    & \multicolumn{2}{c}{Prefill} & \multicolumn{2}{c}{Train} & \multicolumn{2}{c}{Prefill} & \multicolumn{2}{c}{Train} \\
    \cmidrule(lr){2-3}\cmidrule(lr){4-5}\cmidrule(lr){6-7}\cmidrule(l){8-9}
    Model & ms & GiB & ms & GiB & ms & GiB & ms & GiB \\
    \midrule
    \multicolumn{9}{@{}l}{\textit{1 GPU}} \\
    Llama & 56.8 & 2.27 & 179.5 & 9.52 & 274.3 & 13.96 & 906.0 & 54.72 \\
    mHC4 & 68.4 & 2.32 & 226.4 & 12.44 & 301.8 & 14.19 & 1009.4 & 58.99 \\
    \method{}4 & 73.4 & 2.45 & 243.4 & 15.93 & 314.9 & 14.44 & 1093.5 & 69.11 \\
    \method{}8 & 83.7 & 2.73 & 291.8 & 23.06 & 344.5 & 15.04 & 1234.5 & 89.72 \\
    \method{}16 & 90.9 & 3.16 & 325.4 & 29.34 & 363.8 & 16.00 & \multicolumn{2}{c}{OOM} \\
    \midrule
    \multicolumn{9}{@{}l}{\textit{4 GPUs, TP degree 4}} \\
    Llama & 25.1 & 0.90 & 80.0 & 3.90 & 87.2 & 3.94 & 285.8 & 18.23 \\
    mHC4 & 40.9 & 0.91 & 156.0 & 6.72 & 116.5 & 4.01 & 424.7 & 26.55 \\
    \quad + sharded streams & 36.8 & 0.94 & 151.8 & 4.83 & 112.4 & 3.96 & 411.5 & 20.66 \\
    \method{}4 & 44.7 & 0.95 & 185.8 & 10.04 & 137.2 & 4.23 & 463.2 & 33.79 \\
    \quad + sharded streams & 37.8 & 0.94 & 193.3 & 5.37 & 118.6 & 4.01 & 447.0 & 21.73 \\
    \method{}8 & 54.8 & 1.21 & 216.1 & 17.03 & 167.1 & 4.76 & 624.9 & 54.45 \\
    \quad + sharded streams & 42.5 & 0.97 & 199.9 & 6.13 & 122.1 & 4.15 & 494.4 & 23.96 \\
    \method{}16 & 61.1 & 1.65 & 256.3 & 23.36 & 191.0 & 5.77 & 735.4 & 72.39 \\
    \quad + sharded streams & 45.9 & 1.10 & 206.2 & 7.37 & 133.1 & 4.54 & 531.0 & 27.40 \\
    \bottomrule
  \end{tabular}
\caption{Prefill and training cost on NVIDIA H20 GPUs with batch size 1 and sequence length 2048. We report the average prefill latency and training step time (ms), together with the peak memory per GPU (GiB). Under TP, the streams are either replicated on every GPU, as in Megatron TP, or sharded across GPUs with ``+ sharded streams'' (\S\ref{sec:tp}). 
Training is compiled with \texttt{max-autotune}, and prefill uses the same mode with CUDA graphs disabled.}
\label{tab:tpbench}
\end{table*}

\subsection{Efficiency}
\label{sec:efficiency}


We test the prefill and training cost of the TinyLlama-1.1B and Llama-3.1-8B models \citep{grattafiori2024llama3} under two configurations: single-GPU execution without TP and four-GPU execution with a TP degree of 4.
Table~\ref{tab:tpbench} reports the results.
Without TP at both scales, \method{}4 introduces 4--7\% additional prefill latency compared with mHC4, and 7--8\% additional training step time.
Under TP with sharded streams at the 8B scale, \method{}4 is 6\% slower than mHC4 in prefill and 9\% slower in training, while using 5\% more peak training memory. Compared with replicated streams, sharding reduces the prefill latency of \method{}4 by 15\% at 1.1B and 14\% at 8B, versus 10\% and 4\% for mHC4. It also reduces the peak training memory of \method{}4 by 47\% and 36\%, respectively.

At 1.1B, however, sharding increases the training step time of \method{}4 from 185.8 to 193.3~ms. Its training overhead relative to mHC4 is 19\% with replicated streams and 27\% with sharded streams.
A possible reason is that the additional collectives add a roughly fixed cost, which outweighs the savings when the per-GPU workload is small.
Sharding also slightly increases the prefill memory of mHC4 from 0.91 to 0.94 GiB. Apart from these two exceptions, sharding reduces both latency and peak memory.

The benefits of sharding grow with $N$.
At the 8B scale, sharding reduces the prefill latency of \method{}8 and \method{}16 by 27\% and 30\%, their training step time by 21\% and 28\%, and their peak training memory by 56\% and 62\%, respectively.
Scaling from \method{}4 to \method{}16 with sharded streams increases prefill latency by 12\%, training step time by 19\%, and peak training memory by 26\%. The corresponding increases with replicated streams are 39\%, 59\%, and 114\%. On a single GPU, \method{}16 runs out of memory during training.

\subsection{Ablation Studies}
\label{sec:ablation}

\begin{figure}[!t]
  \centering
  \input{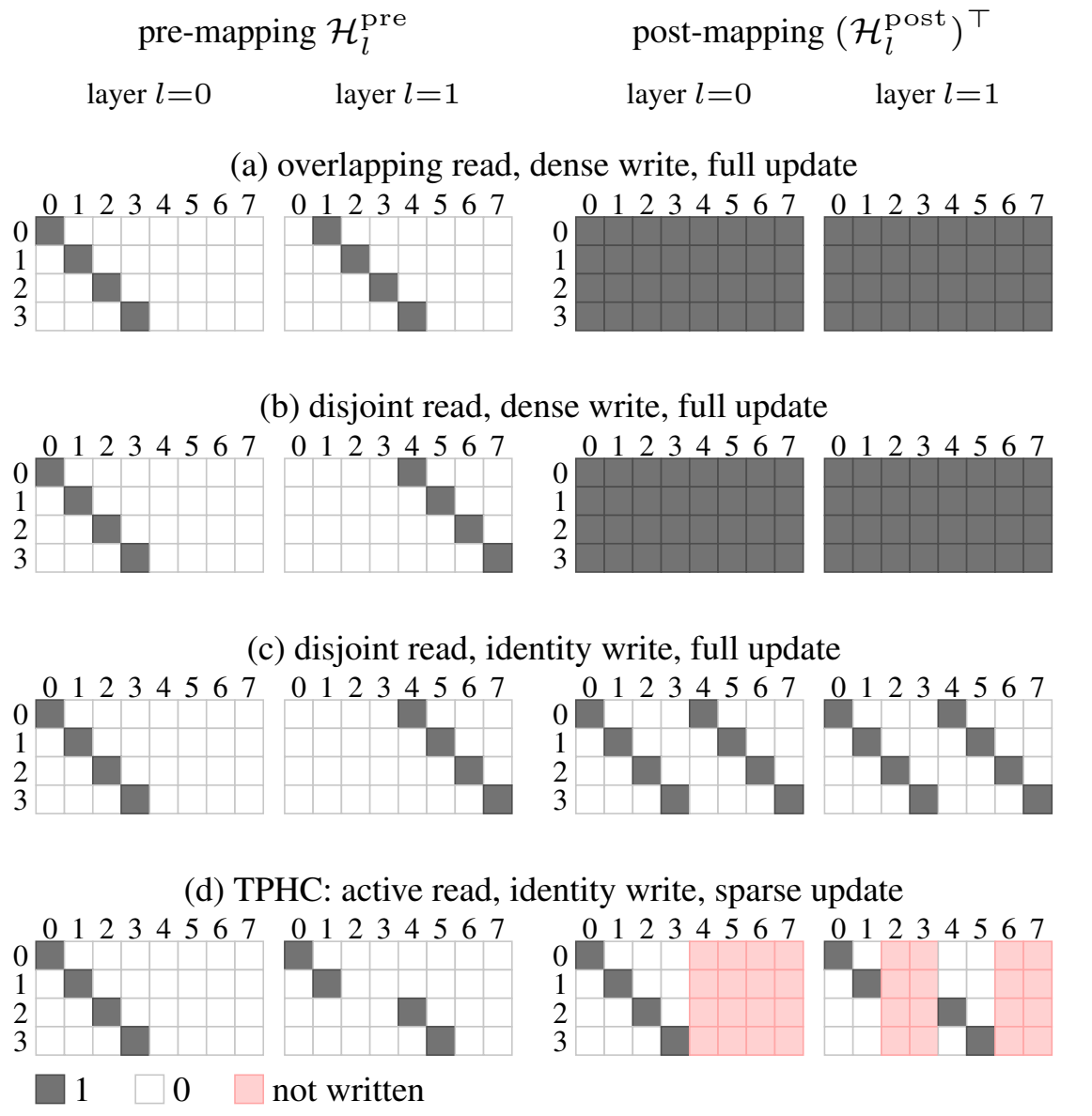}
  \caption{Initial read ($\Hpre_l$) and write ($\Hpost_l$, drawn transposed) patterns of the configurations in Table~\ref{tab:abl} for $N{=}8$ streams and $k{=}4$ groups in the first two layers. Rows are groups, columns are streams, and gray cells are near 1 and white cells near 0. 
  (a)--(c) write all $N$ streams at every layer. (d) writes only the active streams of Eq.~\eqref{eq:schedule}, and the remaining streams are unchanged.}
  \label{fig:init}
\end{figure}

We first evaluate the grouped read and write in isolation and then the three design choices of \method{}: which streams the groups read at initialization, how the write is initialized, and how many streams each layer updates.
All runs train 1.1B models with $k{=}4$ groups for one epoch on MiniPile \citep{kaddour2023minipile}, with a batch of 64K tokens.
We report the final validation loss in Table~\ref{tab:abl}, and show the initial mappings in Figure~\ref{fig:init}.

\begin{table}[t]
  \centering
  \small
  \setlength{\tabcolsep}{2.4pt}
  \renewcommand{\arraystretch}{1.15}
  \begin{tabular}{@{}lllcccc@{}}
    \toprule
    & \multicolumn{2}{c}{Initialization} & & \multicolumn{3}{c}{Loss} \\
    \cmidrule(lr){2-3}\cmidrule(l){5-7}
    & read & write & update & $N{=}4$ & $N{=}8$ & $N{=}16$ \\
    \midrule
    mHC & \multicolumn{2}{l}{no groups} & full & 2.2377 & 2.2271 & 2.2281 \\
    \midrule
    (a) & overlap & dense & full & \multirow{2}{*}{\textbf{2.2324}} & 2.2311 & 2.2202 \\
    (b) & disjoint & dense & full & & 2.2250 & 2.2203 \\
    (c) & disjoint & identity & full & \multirow{2}{*}{2.2328} & 2.2213 & \textbf{2.2095} \\
    (d) \method{} & active & identity & sparse & & \textbf{2.2195} & 2.2100 \\
    \bottomrule
  \end{tabular}
  \caption{Ablation of \method{} with $k{=}4$ groups. The final validation loss is reported. Lower is better. Rows (a)--(d) correspond to Figure~\ref{fig:init}(a)--(d),
  and (d) is the final \method{}. mHC uses the same number of streams without groups. With $N{=}k{=}4$, every layer reads and writes all streams, so (a) and (b) share one run, as do (c) and (d).}
  \label{tab:abl}
\end{table}

\paragraph{Grouped read and write.}

Row (a) of Table~\ref{tab:abl} adds the grouped read and write to mHC but retains its initialization: $\Hpost_l$ starts as an all-ones matrix, so all streams are initially identical (\emph{dense write}), and every layer writes all $N$ streams.
The grouped read and write (a) reduce the loss of mHC by 0.005 at $N{=}4$ and 0.008 at $N{=}16$ but increase it by 0.004 at $N{=}8$, which motivates the following changes.

\paragraph{Read.}
In (a), the streams read at initialization shift by one from layer to layer, so consecutive layers read largely the same streams.
Disjoint read (b) gives consecutive layers disjoint inputs and covers all $N$ streams within $N/k$ layers, reducing the loss by 0.006 at $N{=}8$ and leaves it unchanged at $N{=}16$, and we therefore adopt it.

\paragraph{Write.}
In (b), dense write initialization produces the same write-back component across streams, whereas the identity write in (c) gives each stream a distinct one.
With 4 streams, the two settings differ in validation loss by less than 0.001, whereas with 8 and 16 streams (c) reduces the loss by 0.004 and 0.011, respectively, and we therefore adopt it.

\paragraph{Update.}
In the sparse update (d), each group initially reads from and writes to its own active stream, which significantly reduces the computational cost of mapping generation.
Its loss remains within 0.002 of the full update (c) at both $N{=}8$ and $16$, and we therefore adopt it.

\begin{figure}[t]
  \centering
\begin{tikzpicture}
\pgfplotsset{
  div/.style={scale only axis, width=0.87\columnwidth, height=1.85cm,
    xmin=0, xmax=45, xtick={1,10,20,30,40,44}, xlabel={Sublayer}, enlarge y limits=0.06,
    tick label style={font=\scriptsize}, label style={font=\scriptsize},
    tick align=outside, tick pos=left,
    axis line style={black!60}, major tick length=2pt,
    grid=major, grid style={black!8}, title style={font=\scriptsize, yshift=-5pt}},
  smhc/.style={draw=mhcc, line width=0.75pt, mark=*, mark size=0.8pt, mark options={fill=mhcc, draw=mhcc}},
  sxhc/.style={draw=xhcc, line width=0.75pt, mark=triangle*, mark size=1.0pt, mark options={fill=xhcc, draw=xhcc}},
  stphc/.style={draw=tphcc, line width=0.75pt, mark=square*, mark size=0.8pt, mark options={fill=tphcc, draw=tphcc}},
  ref/.style={black!45, densely dashed, line width=0.6pt},
}
\begin{axis}[div, name=a, title={(a) Mean pairwise cosine similarity between streams},
  ymin=0.3, ymax=1.02, ytick={0.4,0.6,0.8,1.0}, enlarge y limits=false,
  legend style={font=\scriptsize, draw=none, fill=white, fill opacity=0.85, text opacity=1,
    at={(0.98,0.04)}, anchor=south east, legend columns=3, column sep=5pt}]
\addplot[smhc] coordinates {(1,0.6016) (2,0.7231) (3,0.6688) (4,0.6745) (5,0.7282) (6,0.7802) (7,0.7978) (8,0.8131) (9,0.7870) (10,0.7721) (11,0.8331) (12,0.8571) (13,0.8043) (14,0.7945) (15,0.8090) (16,0.7927) (17,0.7899) (18,0.7912) (19,0.8263) (20,0.8184) (21,0.8021) (22,0.8008) (23,0.8146) (24,0.8174) (25,0.8244) (26,0.8187) (27,0.8263) (28,0.8295) (29,0.8624) (30,0.8511) (31,0.8913) (32,0.8840) (33,0.8992) (34,0.8947) (35,0.8965) (36,0.8987) (37,0.9067) (38,0.9016) (39,0.9060) (40,0.9064) (41,0.9531) (42,0.9550) (43,0.9725) (44,0.9848)};
\addlegendentry{mHC4}
\addplot[sxhc] coordinates {(1,0.6057) (2,0.3622) (3,0.3880) (4,0.5934) (5,0.5726) (6,0.6523) (7,0.6447) (8,0.7777) (9,0.7947) (10,0.7439) (11,0.7021) (12,0.7453) (13,0.7420) (14,0.7341) (15,0.7767) (16,0.7893) (17,0.6779) (18,0.6920) (19,0.7038) (20,0.6881) (21,0.7097) (22,0.7210) (23,0.7265) (24,0.7219) (25,0.7216) (26,0.7298) (27,0.7577) (28,0.7661) (29,0.6979) (30,0.7182) (31,0.7536) (32,0.7699) (33,0.7838) (34,0.8026) (35,0.8059) (36,0.8207) (37,0.8295) (38,0.8452) (39,0.8581) (40,0.8654) (41,0.8923) (42,0.9219) (43,0.9447) (44,0.9662)};
\addlegendentry{mHC-TFA4}
\addplot[stphc] coordinates {(1,0.5849) (2,0.5761) (3,0.6075) (4,0.5939) (5,0.6181) (6,0.5978) (7,0.6358) (8,0.6193) (9,0.6518) (10,0.6596) (11,0.6837) (12,0.6903) (13,0.6986) (14,0.6829) (15,0.6889) (16,0.6723) (17,0.7062) (18,0.6883) (19,0.7013) (20,0.6958) (21,0.7157) (22,0.7095) (23,0.7172) (24,0.7034) (25,0.6969) (26,0.6796) (27,0.6944) (28,0.6792) (29,0.6814) (30,0.6766) (31,0.6927) (32,0.6835) (33,0.7010) (34,0.6983) (35,0.7134) (36,0.7372) (37,0.7512) (38,0.7687) (39,0.7988) (40,0.8107) (41,0.8510) (42,0.8692) (43,0.9048) (44,0.9173)};
\addlegendentry{\method{}4}
\end{axis}
\begin{axis}[div, name=b, at={(a.south west)}, anchor=north west, yshift=-1.3cm,
  title={(b) Mean pairwise cosine similarity between rows of $\mathcal{H}^{\mathrm{post}}_l$},
  ymin=0, ymax=1.05, ytick={0,0.25,0.5,0.75,1.0}, enlarge y limits=false]
\addplot[smhc] coordinates {(1,1.0000) (2,1.0000) (3,1.0000) (4,1.0000) (5,1.0000) (6,1.0000) (7,1.0000) (8,1.0000) (9,1.0000) (10,1.0000) (11,1.0000) (12,1.0000) (13,1.0000) (14,1.0000) (15,1.0000) (16,1.0000) (17,1.0000) (18,1.0000) (19,1.0000) (20,1.0000) (21,1.0000) (22,1.0000) (23,1.0000) (24,1.0000) (25,1.0000) (26,1.0000) (27,1.0000) (28,1.0000) (29,1.0000) (30,1.0000) (31,1.0000) (32,1.0000) (33,1.0000) (34,1.0000) (35,1.0000) (36,1.0000) (37,1.0000) (38,1.0000) (39,1.0000) (40,1.0000) (41,1.0000) (42,1.0000) (43,1.0000) (44,1.0000)};
\addplot[sxhc] coordinates {(1,1.0000) (2,0.6675) (3,1.0000) (4,0.7744) (5,1.0000) (6,0.5833) (7,1.0000) (8,0.8599) (9,1.0000) (10,0.6866) (11,1.0000) (12,0.6850) (13,1.0000) (14,0.6628) (15,1.0000) (16,0.8763) (17,1.0000) (18,0.8160) (19,1.0000) (20,0.8688) (21,1.0000) (22,0.7094) (23,1.0000) (24,0.6935) (25,1.0000) (26,0.7381) (27,1.0000) (28,0.8886) (29,1.0000) (30,0.8777) (31,1.0000) (32,0.9233) (33,1.0000) (34,0.9193) (35,1.0000) (36,0.8509) (37,1.0000) (38,0.7580) (39,1.0000) (40,0.7301) (41,1.0000) (42,0.8677) (43,1.0000) (44,1.0000)};
\addplot[stphc] coordinates {(1,0.0394) (2,0.1219) (3,0.4829) (4,0.1896) (5,0.2964) (6,0.1629) (7,0.3337) (8,0.1850) (9,0.4060) (10,0.2179) (11,0.3541) (12,0.1780) (13,0.3528) (14,0.1179) (15,0.4024) (16,0.1616) (17,0.3459) (18,0.1661) (19,0.4920) (20,0.2265) (21,0.6491) (22,0.2385) (23,0.2905) (24,0.2321) (25,0.3312) (26,0.2585) (27,0.6094) (28,0.2925) (29,0.4839) (30,0.3393) (31,0.4868) (32,0.3906) (33,0.6156) (34,0.4611) (35,0.5641) (36,0.5903) (37,0.6415) (38,0.7002) (39,0.5390) (40,0.7451) (41,0.7453) (42,0.7956) (43,0.7587) (44,0.7512)};
\end{axis}
\end{tikzpicture}
  \caption{Stream collapse and writes across depth for mHC4, mHC-TFA4 and \method{}4. The 44 sublayers are numbered in order, with attention at odd and FFN at even positions. (a) Mean pairwise cosine similarity between streams after each sublayer, averaged over tokens. (b) Mean pairwise cosine similarity between the rows of the token-averaged write matrix $\Hpost_l$ of each sublayer,
  which is 1 whenever $\Hpost_l$ is a single column, as in mHC and in the attention sublayers of mHC-TFA4.}
  \label{fig:cos}
\end{figure}
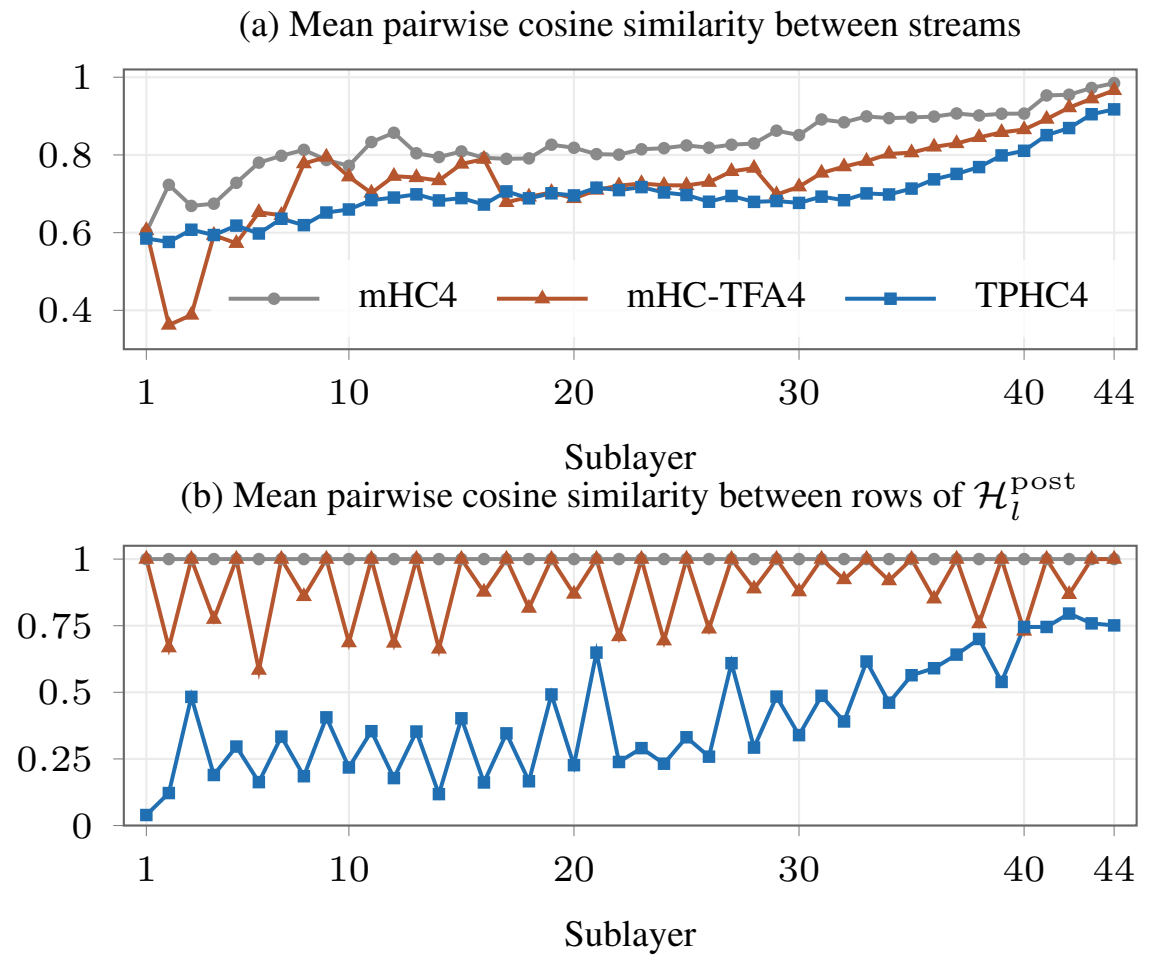

With all three choices, \method{} improves over mHC by 0.005, 0.008 and 0.018 at $N{=}4$, $8$ and $16$, and unlike mHC, it continues to improve from 8 to 16 streams.

\subsection{Stream Diversity Analysis}
\label{sec:diversity}

To examine how the models use their streams, we measure the mean pairwise cosine similarity between streams after each sublayer on SlimPajama validation data (Figure~\ref{fig:cos}a).
In mHC4, the streams become increasingly aligned with depth, reaching a cosine similarity of 0.985 at the last layer, while their norms remain comparable.
The multi-stream residual thus collapses toward a single direction, even though all streams remain in use.

A likely cause is mHC's shared write-back.
$\Hres_l$ remains close to the identity in most layers, and the stream norms grow by more than an order of magnitude with depth, so each stream is dominated by its accumulated writes.
In mHC, these writes are scalar multiples of the same sublayer outputs, so all streams are weighted sums of the same vectors and differ only in their weights.

mHC-TFA4 and \method{}4 both write multiple distinct outputs, the four TFA components of the FFN output and the group outputs, respectively.
Both reduce the stream similarity relative to mHC4, and \method{}4 attains the lowest similarity throughout the second half of the network, ending at 0.917 compared with 0.966 for mHC-TFA4.
The two differ in the similarity of the write mixtures that different streams receive (Figure~\ref{fig:cos}b).
In mHC-TFA4, the attention sublayers still write a single output, since TFA augments only the FFN, and even in the FFN sublayers the streams receive similar mixtures (0.79 on average).
In \method{}4, whose groups are initialized to write to their own streams, the mixtures remain distinct (0.40 on average) and become similar only in the last layers, where the streams also align.

\section{Conclusion}
\label{sec:conclusion}

We propose \method{}, which improves HC by using tensor-parallel sublayer partitions to produce distinct write-back components without increasing the parameters or FLOPs of the attention and FFN sublayers.
\method{} achieves lower training and validation loss than mHC and mHC-TFA  for $N\in\{4,8,16\}$, raises the average zero-shot commonsense reasoning accuracy of mHC, and with 8 streams outperforms mHC with 16.
We also implement TP with sharded streams for \method{}, keeping the training-time overhead of \method{}4 relative to mHC4 small at the 8B scale and makes \method{}16 affordable.
Our analysis further shows that \method{}4 maintains lower inter-stream cosine similarity than mHC4 and mHC-TFA4, consistent with its goal of preserving stream diversity.

\section*{Limitations}

Our pretraining experiments are limited to 1.1B-parameter models and 100B tokens, and thus do not cover larger models with more training data.
Our work focuses on dense Transformers, where \method{} builds its groups around tensor-parallel partitions.
Consequently, the current grouping and communication design cannot be directly transferred to MoE models with expert parallel.
We leave the extension of \method{} to MoE models for future work.

\section*{Ethics Statement}
This work proposes an architectural modification for language-model pre-training and inherits the risks of the underlying models and training data. It does not introduce new data collection or new capabilities beyond those of standard LLMs. All models are trained on publicly released corpora, SlimPajama \citep{soboleva2023slimpajama} and, for the ablations, MiniPile \citep{kaddour2023minipile}, and inherit their biases. SlimPajama is released under the Apache 2.0 license and MiniPile under the MIT license of the Pile, both of which permit research use.

\section*{Acknowledgements}
We would like to thank Xuhao Lu for insightful suggestions on this work.

\bibliography{custom}

\appendix
\raggedbottom

\section{Implementation Details}
\label{app:impl}

\paragraph{Batched group computation.}
All per-group projections are stored as tensors with a leading group dimension, $W\in\R^{k\times d_{\mathrm{out}}\times d_{\mathrm{in}}}$, and applied with a single batched matrix multiplication over the $k$ groups. The query, key and value projections are fused into one batched GEMM, and so are the SwiGLU gate and up projections. After the projection, the heads of all groups are concatenated so that a standard attention kernel (e.g., SDPA) computes attention over all $n_h$ heads at once. Group $j$'s queries, keys and values are computed from its own normalized input $\mathrm{RMSNorm}(\mathbf{u}_j)$.

\paragraph{Functional scatter.}
The write of Eqs.~\eqref{eq:write}--\eqref{eq:identity} is implemented as $X_l=X_{l-1}\odot(1-\mathbf{a}_l)+P_l\,\big(\Hres_l X_{l-1}[\Sset_l]+\Hpost_l O_l\big)$, where $P_l\in\{0,1\}^{N\times k}$ is the one-hot scatter matrix of $\Sset_l$ and $\mathbf{a}_l=P_l\mathbf{1}_k$ is the active-stream mask. Both are fixed buffers, which avoids in-place indexing operations that break compiled autograd. This functional form reads and writes the full residual state, whereas the memory traffic in Table~\ref{tab:mem} counts an in-place scatter that touches only the active streams.

\paragraph{Initialization.}
With $\boldsymbol{\varphi}=\mathbf{0}$ and $\alpha=0.01$, each mapping starts at its static bias. \method{} uses bias entries $\pm\beta$ with $\beta{=}3$.
The identity write of \method{} sets $b^{\mathrm{pre}}_{ji}=+\beta$ if $i=s_j$ and $\mathbf{b}^{\mathrm{post}}=\mathbf{b}^{\mathrm{res}}=\beta(2\mathbf{I}-\mathbf{1}\mathbf{1}^{\top})$, where $\mathbf{1}$ is the all-ones vector, with all other entries $-\beta$.

\paragraph{Full-update variant.}
The configurations of Figure~\ref{fig:init}a--c use the full update of Eq.~\eqref{eq:fullwrite}, with all mappings generated from the $NC$-dimensional state. In (a), $b^{\mathrm{pre}}_{ji}=+\beta$ if $i=(l+j)\bmod N$, so consecutive layers read overlapping streams. In (b) and (c), $b^{\mathrm{pre}}_{ji}=+\beta$ if $i=(lk+j)\bmod N$, so consecutive layers read disjoint streams. The dense write of (a) and (b) computes $\Hpost_l$ with $2\sigma$ and $\mathbf{b}^{\mathrm{post}}=\mathbf{0}$, so all its entries are 1. The identity write of (c) uses $\sigma$ and sets $b^{\mathrm{post}}_{ij}=+\beta$ if $j=i\bmod k$. In all three, $\mathbf{b}^{\mathrm{res}}=\beta(2\mathbf{I}-\mathbf{1}\mathbf{1}^{\top})$, so $\Hres_l$ starts near the identity, and all other bias entries are $-\beta$.

\paragraph{Numerics.}
Sinkhorn--Knopp runs in fp32 for 20 iterations, alternately normalizing columns and rows after subtracting the column-wise maximum of the logits for stability.

\section{Update Schedule Details}
\label{app:schedule}

\begin{figure}[H]
  \centering
  \resizebox{\columnwidth}{!}{\input{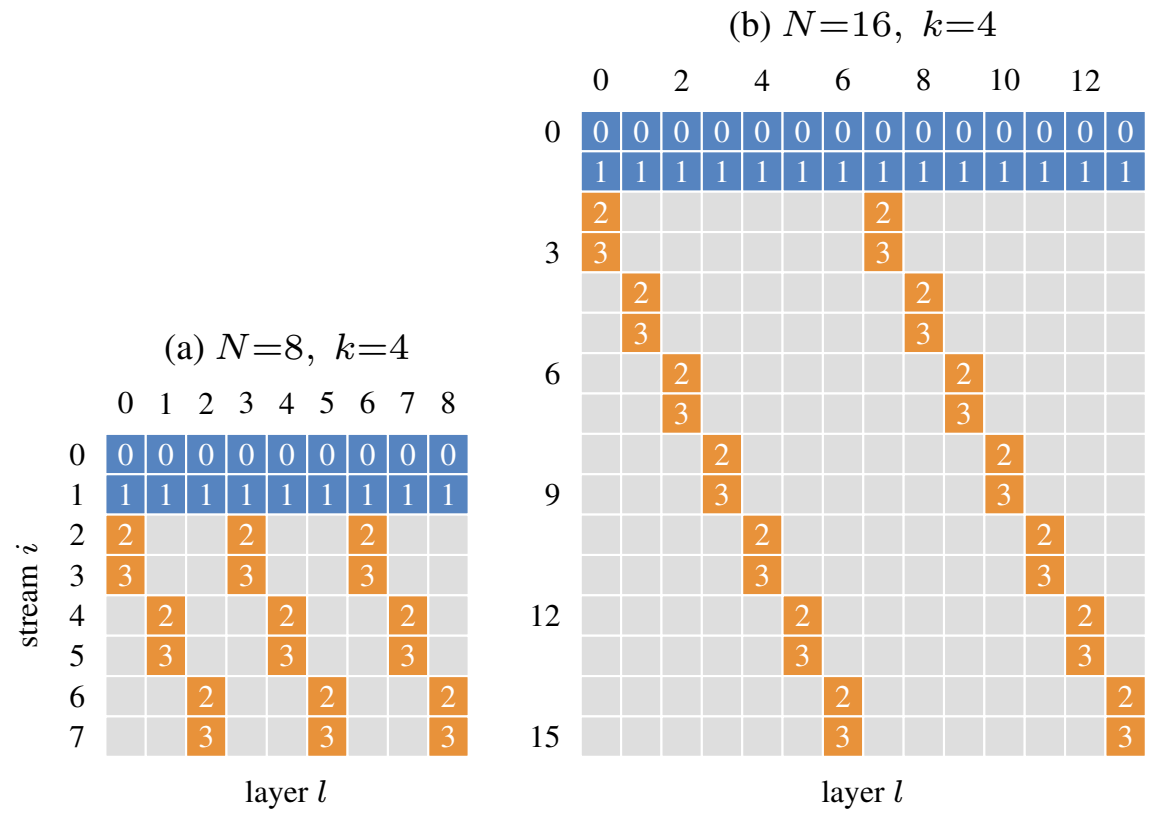}}
  \caption{Active streams of each layer $l$ produced by Eq.~\eqref{eq:schedule}. Blue: fixed streams, written at every layer. Orange: sliding streams. Grey: not written (identity). The number in each active cell is the group $j$ that writes it. Every stream is read at every sublayer.}
  \label{fig:schedule}
\end{figure}

Figure~\ref{fig:schedule} shows the schedule for $k{=}4$. Groups 0 and 1 always write the two fixed streams, and groups 2 and 3 move over the sliding streams, so the pattern repeats every 3 layers for $N{=}8$ and every 7 layers for $N{=}16$. In our 22-layer model with $N{=}16$, the two fixed streams are written in all 22 layers and each of the 14 sliding streams in 3--4 layers, while every stream is read by all 44 sublayers.

\section{Training FLOPs and Parameter Overhead}
\label{app:cost}

We follow the analysis of \citet{zhang2026xhc} and count the mapping generators of one layer (an attention and an FFN sublayer), ignoring biases and gates.

\subsection{Parameters per Layer}
In mHC, each sublayer projects the $NC$-dimensional state to $N^2{+}2N$ coefficients, so a layer adds $P_{\mathrm{mHC}}=(4N^2+2N^3)C$ parameters.
\method{} generates $\Hpre_l$ from the full state ($kN^2C$) and $\Hpost_l$ and $\Hres_l$ from the active streams ($k^3C$ each), so $P_{\mathrm{TPHC}}=(2kN^2+4k^3)C$.

\subsection{Training FLOPs}
As in xHC, the parametric training FLOPs of the mappings are $F_{\mathrm{HC}}=6P_{\mathrm{HC}}L$ per token for $L$ layers; Sinkhorn normalization and the stream mixing itself are left out and appear as memory traffic in Table~\ref{tab:mem}.
The backbone costs six times its linear-layer parameters per token, plus $6Sn_hd_hL$ for causal attention over sequence length $S$ and $6VC$ for the LM head.
Table~\ref{tab:cost-ill} gives both overheads for our 1.1B models ($C{=}2048$, $L{=}22$, $S{=}2048$, $n_h{=}32$, $d_h{=}64$, $V{=}32000$).

\begin{table}[H]
  \centering
  \small
  \setlength{\tabcolsep}{5pt}
  \begin{tabular}{@{}lccc@{}}
    \toprule
    & \multicolumn{3}{c}{1.1B ($C{=}2048$, $L{=}22$)} \\
    \cmidrule(l){2-4}
    Method & $P$/layer & $\Delta P$ & $\Delta F$ \\
    \midrule
    mHC4 & 393K ($192C$) & 0.8\% & 0.8\% \\
    \method{}4 & 786K ($384C$) & 1.6\% & 1.5\% \\
    \midrule
    mHC8 & 2.62M ($1280C$) & 5.2\% & 5.1\% \\
    \method{}8 & 1.57M ($768C$) & 3.1\% & 3.1\% \\
    \midrule
    mHC16 & 18.9M ($9216C$) & 37.7\% & 36.9\% \\
    \method{}16 & 4.72M ($2304C$) & 9.4\% & 9.2\% \\
    \bottomrule
  \end{tabular}
  \caption{Parameters per layer added by the mappings, and the resulting parameter ($\Delta P$) and training-FLOPs ($\Delta F$) overhead relative to the 1.1B backbone, computed as in \citet{zhang2026xhc}. \method{} uses $k{=}4$ and, for $N{>}4$, the sparse update.}
  \label{tab:cost-ill}
\end{table}

\subsection{Memory Traffic}
We follow \citet{zhang2026xhc} in counting memory accesses for the residual-stream operations. Table~\ref{tab:mem} reports the resulting memory traffic and the per-device communication volume under TP, both per token per sublayer.
\begin{table}[H]
  \centering
  \footnotesize
  \setlength{\tabcolsep}{3pt}
  \begin{tabular}{@{}lcccc@{}}
    \toprule
    & \multicolumn{2}{c}{mHC} & \multicolumn{2}{c}{\method{}} \\
    \cmidrule(lr){2-3}\cmidrule(l){4-5}
    Operation & R & W & R & W \\
    \midrule
    Mappings & $NC$ & $N^2{+}2N$ & $NC$ & $kN$ \\
    $\Hpre$ dense read & $NC{+}N$ & $C$ & $NC{+}kN$ & $kC$ \\
    Gather & -- & -- & $kC$ & $kC$ \\
    Active mappings & -- & -- & $kC$ & $2k^2$ \\
    $\Hres$ mixing & $NC{+}N^2$ & $NC$ & $kC{+}k^2$ & $kC$ \\
    $\Hpost$ write-back & $C{+}N$ & $NC$ & $kC{+}k^2$ & $kC$ \\
    Merge and scatter & $2NC$ & $NC$ & $2kC$ & $kC$ \\
    \midrule
    Sum, $N{=}16$ & $81C$ & $49C$ & $56C$ & $20C$ \\
    \midrule
    Total, $N{=}16$ & \multicolumn{2}{c}{$130C$} & \multicolumn{2}{c}{$76C$} \\
    Total, $N{=}8$ & \multicolumn{2}{c}{$66C$} & \multicolumn{2}{c}{$60C$} \\
    Total, $N{=}4$ & \multicolumn{2}{c}{$34C$} & \multicolumn{2}{c}{$44C^{\dagger}$} \\
    \midrule
TP, replicated
  & \multicolumn{2}{c}{$3C/2$}
  & \multicolumn{2}{c}{$3kC/4$} \\
TP, sharded
  & \multicolumn{2}{c}{$3(N+2)C/4$}
  & \multicolumn{2}{c}{$3kC/2$} \\
  \bottomrule
  \end{tabular}
  \caption{Per-token memory access of the residual-stream operations of one sublayer, excluding the input and output of the sublayer function, counted as in \citet{zhang2026xhc} with the active streams gathered and scattered in place. R: read, W: write, $k{=}4$ for numerical totals. $^{\dagger}$No gather is needed when $N{=}k$. TP: data each device sends per token and sublayer under TP degree 4, with the streams replicated on every device or sharded across devices (\S\ref{sec:tp}).}
  \label{tab:mem}
\end{table}

\section{Algorithm}
\label{app:alg}

\begin{algorithm}[H]
  \caption{One sublayer (attention or FFN) of layer $l$ in \method{}}
  \label{alg:tphc}
  \small
  \begin{algorithmic}[1]
    \Require state $X\in\R^{N\times C}$; active set $\Sset_l=(s_0,\dots,s_{k-1})$; groups $\mathcal{F}_{l,0..k-1}$
    \State $\mathbf{x}'\gets\mathrm{RMSNorm}(\vecop X)$ \Comment{$\R^{1\times NC}$}
    \State $\mathbf{x}'_{\Sset}\gets$ entries of $\mathbf{x}'$ of the active streams \Comment{$\R^{1\times kC}$}
    \For{$j=0,\dots,k-1$ \textbf{in parallel}} \Comment{group $j$}
      \State $\mathbf{h}^{\mathrm{pre}}_j\gets\sigma(\alpha^{\mathrm{pre}}(\mathbf{x}'\boldsymbol{\varphi}^{\mathrm{pre}}_j)+\mathbf{b}^{\mathrm{pre}}_j)$ \Comment{row $j$ of $\Hpre$}
      \State $\mathbf{u}_j\gets\mathbf{h}^{\mathrm{pre}}_jX$ \Comment{dense read}
      \State $\mathbf{o}_j\gets\mathcal{F}_{l,j}(\mathbf{u}_j)$
    \EndFor
    \State $O\gets(\mathbf{o}_0,\dots,\mathbf{o}_{k-1})^{\top}$ \Comment{outputs kept separate}
    \State $\Hpost\gets\sigma(\alpha^{\mathrm{post}}\mat(\mathbf{x}'_{\Sset}\boldsymbol{\varphi}^{\mathrm{post}})+\mathbf{b}^{\mathrm{post}})$ \Comment{$\R^{k\times k}$}
    \State $\Hres\gets\SK(\exp(\alpha^{\mathrm{res}}\mat(\mathbf{x}'_{\Sset}\boldsymbol{\varphi}^{\mathrm{res}})+\mathbf{b}^{\mathrm{res}}))$ \Comment{$\R^{k\times k}$}
    \State $X[\Sset_l]\gets\Hres X[\Sset_l]+\Hpost O$ \Comment{sparse update}
    \State \Return $X$ \Comment{rows $\notin\Sset_l$ unchanged}
  \end{algorithmic}
\end{algorithm}

\section{Model and Training Configurations}
\label{app:hparams}

\begin{table}[H]
  \centering
  \small
  \setlength{\tabcolsep}{4pt}
  \begin{tabular}{@{}lcc@{}}
    \toprule
    Section & \S\ref{sec:main}--\S\ref{sec:diversity} & \S\ref{sec:efficiency} \\
    \midrule
    Model size & 1.1B & 8B \\
    Layers $L$ & 22 & 32 \\
    Hidden size $C$ & 2048 & 4096 \\
    FFN size $D$ & 5632 & 14336 \\
    Heads $n_h$ / $n_{kv}$ & 32 / 4 & 32 / 8 \\
    Head dimension & 64 & 128 \\
    Vocabulary & 32{,}000 & 32{,}000 \\
    \bottomrule
  \end{tabular}
  \caption{Model configurations.}
  \label{tab:models}
\end{table}

\begin{table}[H]
  \centering
  \small
  \setlength{\tabcolsep}{4pt}
  \begin{tabular}{@{}lcc@{}}
    \toprule
    Section & \S\ref{sec:main}, \S\ref{sec:diversity} & \S\ref{sec:ablation} \\
    \midrule
    Training data & SlimPajama & MiniPile \\
    Training tokens & 100B & 1.7B \\
    Global batch (tokens) & 2M & 64K \\
    Sequence length & \multicolumn{2}{c}{2048} \\
    LR schedule & \multicolumn{2}{c}{cosine} \\
    Optimizer & \multicolumn{2}{c}{AdamW} \\
    $\beta_1$ & \multicolumn{2}{c}{0.9} \\
    $\beta_2$ & 0.95 & 0.999 \\
    Warmup & 200 steps & 1.5\% of steps \\
    Weight decay / grad.\ clip & \multicolumn{2}{c}{0.1 / 1.0} \\
    Max LR & \multicolumn{2}{c}{$4\times10^{-4}$} \\
    Min LR & $4\times10^{-5}$ & 0 \\

    Precision & \multicolumn{2}{c}{bf16} \\
    GPU & 128 $\times$ H800 & 8 $\times$ H800 \\
    Sinkhorn iterations & \multicolumn{2}{c}{20} \\
    Gate init $\alpha$ / bias margin $\beta$ & \multicolumn{2}{c}{0.01 / 3} \\
    \bottomrule
  \end{tabular}
  \caption{Training configurations of the 1.1B models.
  }
  \label{tab:hparams}
\end{table}

\end{document}